\documentclass[11pt, a4paper,  notitlepage]{extarticle}

\usepackage[left=3.5cm, right=2.5cm, top=3cm, bottom=3cm, bindingoffset=0cm]{geometry}
\usepackage{graphicx}
\usepackage{mathtools}
\usepackage[affil-it]{authblk}
\usepackage{url}
\usepackage{amsmath}
\usepackage{amssymb}
\usepackage{multicol}
\usepackage{float}
\usepackage[section]{placeins}
\usepackage{hyperref}
\usepackage[title]{appendix}

\begin{document}
\sloppy
\title{Using GPT Models for Qualitative and Quantitative News Analytics in the 2024 US Presidential Election Process}
\author{Bohdan M.  Pavlyshenko \\  \small{b.pavlyshenko@gmail.com,  www.linkedin.com/in/bpavlyshenko/ }}
\maketitle

\begin{abstract}
The paper considers an approach of using Google Search API and GPT-4o model for qualitative and quantitative analyses of news through retrieval-augmented generation (RAG). This approach was applied to analyze news about the 2024 US presidential election process. Different news sources for different time periods have been analyzed. Quantitative scores generated by GPT model have been analyzed using Bayesian regression to derive trend lines. The distributions found for the regression parameters allow for the analysis of uncertainty in the election process. The obtained results demonstrate that using the GPT models for news analysis,  one can get informative  analytics and provide key insights that can be applied in further analyses of election processes. 

\end{abstract}

\tableofcontents

\section{Introduction}
Generative Pre-trained (GPT) Models can be effectively used for summarizing and analyzing of different text datasets including news articles.  Large Language Models (LLMs),  due to their transformer structure with attention mechanisms, can help analyze complex texts and reveal different text styles. 
LLMs, such as ChatGPT,  demonstrate high efficiency in the analysis of complex texts~\cite{achiam2023gpt}. 
 GPT models introduce new features compared to conventional transformer-based language models. One of them is zero-shot and few-shot learning, where 
 the model performs well with only a few training examples or even no examples at all, relying solely on the instructions describing what should be done. Another important feature is reasoning, where a model can generate new patterns and conclusions  based on an input prompt and the facts known by the model  which were not included into it directly during the training process.  One of the approaches of using LLMs is based on retrieval-augmented generation (RAG), which uses the results from other services, e.g. relational databases, semantic search, graph database, in the input prompt for the LLM. In this case,   the response  can be treated as the combination of  external results and the LLM's internal knowledge.  

In~\cite{pavlyshenko2023financial},  we use the  fine-tuned Llama 2 LLM model  model~\cite{touvron2023llama}  for financial news analytics.
We study the possibility of fine-tuning Llama 2 Large Language Model (LLM) for the multitask analysis of financial news. For fine-tuning, the PEFT/LoRA based approach was used. 
The obtained results show that the fine-tuned Llama 2 model can perform a multitask financial news analysis with a specified structure of response, part of response can be a structured text and another part of data can have JSON format  for further processing. 
 In~\cite{pavlyshenko2023analysis} we consider the possibility of fine-tuning the Llama 2 LLM for the disinformation analysis and fake news detection.   In the study, the model was fine-tuned for the following tasks: analysing a text to identify disinformation and  propaganda narratives, fact checking, fake news detection,  manipulation analytics, extracting named entities along with their sentiments. The obtained results show that the fine-tuned Llama 2 model can conduct a deep analysis of texts and reveal complex styles and narratives.  The extracted sentiments for named entities can be considered as predictive features in supervised machine learning models. 
The obtained results indicate that the fine-tuned Llama 2 model can perform multitask news analyses with a specified structure of response, where one part of it can be a structured text and another part of the data can have a JSON format that is convenient for further processing of LLM response. 
The fine-tuned Llama 2 model was tested on different news data including theses, statements  and claims of well-known people. The results demonstrate 
that a fine-tuned model can conduct inferences and give a chain of thoughts regarding the text under analysis. 
The considered approach shows high efficiency, using small sets of instructions due to the LLM's ability of few-shot learning that is not inherent in conventional transformer based models. This approach can be applied for using extracted entities and sentiments in supervised models with quantitative target variables, e.g. for the analysis of companies' behavior on financial and business markets. In \cite{pavlyshenko2022methods},  different approaches for the analysis of news trends on Twitter were considered. The obtained results show that an effective system for detecting fake and manipulative news can be developed using ca ombined neural network which consists of three concatenated subnetworks. 

Different approaches for text analytics can be found in~\cite{kawintiranon2022polibertweet,argyle2023out,goyal2022news,rodman2024political, luitse2021great,
kheiri2023sentimentgpt,jiang2024disinformation,rozado2023political,
rathje2024gpt,achiam2023gpt, spitale2023ai}.
Paper~\cite{rodman2024political} argues that political theory should shift its perspective on computational methods by integrating generative large language models, such as GPT-4, to support creative theoretical work, while also emphasizing the need to be mindful of the content and value of theorizing, the technical limitations of the models, and the ethical issues they raise.
The work~\cite{jiang2024disinformation} considers three research questions: (1) the effectiveness of current disinformation detection techniques in identifying disinformation generated by LLMs, (2) the potential of using LLMs themselves as a defense against advanced disinformation, and (3) the exploration of novel approaches to counter the threat if traditional methods and LLM-based defenses prove insufficient. A comprehensive investigation into the formation and detection of disinformation is undertaken to support this research.
Work~\cite{kawintiranon2022polibertweet} presents the development of PoliBERTweet, a domain-specific pre-trained language model based on BERTweet, trained on over 83 million US 2020 election-related English tweets, demonstrating that it outperforms general-purpose models in tasks like political misinformation analysis and candidate stance detection, thereby highlighting the importance of specialized resources for social media NLP tasks.
Paper~\cite{argyle2023out} proposes that language models, such as GPT-3, can serve as effective proxies for specific human sub-populations in social science research by demonstrating a property termed "algorithmic fidelity," where the model's biases are fine-grained and demographically correlated, allowing it to emulate diverse human response distributions based on socio-demographic conditioning, thereby offering a nuanced tool to advance interdisciplinary understanding of human attitudes and socio-cultural contexts.
The study~\cite{kheiri2023sentimentgpt} presents  three main strategies: 1) prompt engineering with GPT-3.5 Turbo, 2) fine-tuning GPT models, and 3) an innovative approach to embedding classification. It provides comparative insights into these strategies and individual GPT models, highlighting their strengths and limitations. The study also benchmarks GPT-based approaches against other high-performing models used on the SemEval 2017 dataset, demonstrating that GPT models outperform them by over 22\% in F1-score. Additionally, it addresses challenges like understanding context and sarcasm, showcasing the enhanced capabilities of GPT models in these areas. Overall, the findings emphasize the potential of GPT models in advancing sentiment analysis and suggest directions for future research.
The work~\cite{goyal2022news} considers the comparison between GPT-3 and fine-tuned models on news summarization. It finds that GPT-3, using task description prompts, produces summaries preferred by humans and avoids common dataset-specific issues such as poor factuality. The study also explores the limitations of current evaluation metrics, highlighting that both reference-based and reference-free metrics struggle to accurately assess GPT-3 summaries. Additionally, the research extends to keyword-based summarization, comparing traditional fine-tuning with prompting approaches. To facilitate further research, the authors release a corpus of 10K generated summaries and 1K human preference judgments across multiple summarization benchmarks.
Research~\cite{rathje2024gpt} examines the political biases in Large Language Models (LLMs) by administering 15 political orientation tests to ChatGPT, revealing a consistent preference for left-leaning viewpoints across 14 of the 15 tests, despite the model's claims of neutrality; it argues that ethical AI systems should aim for political neutrality on normative issues, presenting balanced arguments and avoiding misleading assertions of impartiality while displaying bias.
Paper~\cite{rathje2024gpt} reports the development of GPT-4, a large-scale multimodal Transformer-based model capable of processing both image and text inputs to generate text outputs, demonstrating human-level performance on various benchmarks, including scoring in the top 10\% on a simulated bar exam, and achieving improved factuality and behavior alignment through a post-training process, with its performance reliably predicted using models trained on a fraction of the compute.

In this study, we consider an approach using the Google Search API and the GPT-4o model for qualitative and quantitative analysis of news through retrieval-augmented generation (RAG). This approach will be applied to the analysis of news about the  2024 US presidential election. The main goal is not to predict the 2024 US election outcomes or draw any political conclusions. 
Rather, we focus on extracting news information and analyzing it using the GPT model, with the potential for further application at subsequent levels of analytics.

\section{Methodology}
News data for the analysis was received from web resources in two steps:
\begin{enumerate}
\item Retrieiving URLs for relevant web resources using \textit{Google Search API}. For this purpose, Python library \textit{Google API Clent} was used.
\item Extracting information from web resources given their URLs. For this purpose \textit{SeleniumURLLoader} from 
\textit{LangChain} python library was used. 
\end{enumerate}
For searching relevant web resources, the search query \textit{'Kamala Harris AND Donald Trump'} was used, along with search options to specify time periods and news resources. The following news resources were used for separate searches:
\textit{'Web sites', 'The New York Times', 
                            'CNN', 'The Washington Post', 'Fox News', 'NBC News',
                            'Reuters','ABC News','Bloomberg'}. 
                      Web resource \textit{'Web sites'} refers to the top web sites for the search query without specifying a web source. 
We conducted searches for the following time periods: '2024-08-01'--'2024-08-15', '2024-08-16'--'2024-08-31', '2024-09-01'--'2024-09-15', '2024-09-16'--'2024-09-30', '2024-10-01'--'2024-10-15'. 
The URLs of web resources were grouped by time periods and by sources. Then, data from the found URLs were extracted using \textit{SeleniumURLLoader}. In tolal, 436 web resources were loaded. The data from each extracted resource were analyzed using Open AI with GPT-4o model. 
The text data for the analysis were included in a created prompt for the  GPT-4o model, where we specified the options for the analysis, e.g. 
what and how to summarize, which points to highlight, what quantitative scores should be generated. 
We also specified the instructions for the output format which should be in JSON with an appropriate structure. 
The instructions specified the generation of probability scores for candidates to be elected, sentiments scores, and descriptions for each candidate under consideration.
 Main narratives and key points for each candidates were also requested. 
 In response to the OpenAI API, we will receive qualitative and quantitative analytical data  generated by the GPT-4o model, grouped by time periods and web resources.  We also created a prompt for the second level of the analysis using RAG approach. 
 In this prompt, we specified the instructions for the qualitative analysis of data, generated by the GPT-4o model at the first level for separate analysis of each web resource.
 In the prompt instructions, we requested to summarize data grouped by time periods and web resources. 
 Separately, we instructed the model to analyze qualitative trends and to generate sentiment scores.

\section{Qualitative Results}
 The following are the results  of the RAG approach for the qualitative analysis using GPT-4o model grouped by time periods and web resources:
\subsection{Results grouped by time periods}
\subsubsection{Dates: 2024-08-01 -- 2024-08-15}
 \textbf{Summary:} This period in the 2024 US presidential election features Kamala Harris and Donald Trump as the primary candidates. Harris is lauded for her leadership and preventing disruptive chants at rallies, while Trump draws controversy for labeling January 6 as a 'day of love'. Influencers include Bill de Blasio and Charlamagne Tha God, who shape narratives surrounding the candidates' rhetoric and events such as Harris's selection of Tim Walz as running mate gain traction. The rhetoric is polarized, with Harris focusing on justice and equality, and Trump prioritizing economic revival and anti-establishment themes.
\\ \textbf{Harris' probability score:} 0.505
\\ \textbf{Trump's probability score:} 0.495
\\ \textbf{Harris' positive sentiments:} Presidential demeanor; rule-of-law respect; Democratic support.
\\ \textbf{Trump's positive sentiments:} Strong base support; charismatic leadership.
\\ \textbf{Harris' negative sentiments:} Policy inconsistencies; aggressive exchanges during protests.
\\ \textbf{Trump's negative sentiments:} Controversial stances on January 6 and personal attacks.
\\ \textbf{Harris' cites:} Democrats believe in the rule of law.
\\ \textbf{Trump's cites:} January 6 was a day of love.
\\ \textbf{Harris' main narratives:} Focus on legality and justice, projecting a presidential image.
\\ \textbf{Trump's main narratives:} Maintains influence through controversial and populist rhetoric.
\\ \textbf{Favorite candidate summary:} Harris slightly favored due to broader voter appeal in key demographics.
\subsubsection{Dates: 2024-08-16 -- 2024-08-31}
 \textbf{Summary:} During this period, Kamala Harris and Donald Trump engage in competitive campaigning as the 2024 US presidential election approaches. Harris gains traction with demographic support although Trump earns enthusiasm from conservatives. Debate preparation highlights both candidates' readiness to address economic issues and immigration. Key figures like Musk influence Trump's strategy, while Harris focuses on moderate policies and union appeal. Swing states remain tightly contested.
\\ \textbf{Harris' probability score:} 0.52
\\ \textbf{Trump's probability score:} 0.48
\\ \textbf{Harris' positive sentiments:} Commitment to economic reform, positive Democratic momentum.
\\ \textbf{Trump's positive sentiments:} Strong base support, effective low-propensity voter reach.
\\ \textbf{Harris' negative sentiments:} Economic and policy criticism from moderates.
\\ \textbf{Trump's negative sentiments:} Controversies related to rhetoric and perceived instability.
\\ \textbf{Harris' cites:} Commitments to family economic support.
\\ \textbf{Trump's cites:} Past achievements, assurances from platforms like X.
\\ \textbf{Harris' main narratives:} Focus on moderate governance and economic policies.
\\ \textbf{Trump's main narratives:} Economic revival, populism, and social critiques.
\\ \textbf{Favorite candidate summary:} Harris slightly ahead due to focused moderate appeal and successful engagement with younger demographics.
\subsubsection{Dates: 2024-09-01 -- 2024-09-15}
 \textbf{Summary:} Kamala Harris inches ahead in the 2024 presidential race amid strong debate performances and strategic endgame moves. Endorsements from Taylor Swift and discussions around reproductive rights bolster Harris's appeal. Trump's campaign gears up with core conservative outreach and critique of U.S. foreign policy. Swing state dynamics, particularly in the midwest, will heavily influence ultimate electoral outcomes. Although both candidates face narratives of change and continuity from their party bases, Harris capitalizes on broader social justice themes.
\\ \textbf{Harris' probability score:} 0.525
\\ \textbf{Trump's probability score:} 0.475
\\ \textbf{Harris' positive sentiments:} Endorsements, growth in swing state appeal.
\\ \textbf{Trump's positive sentiments:} Charisma, experienced campaign trail presence.
\\ \textbf{Harris' negative sentiments:} Concerns over strategic plans and viability.
\\ \textbf{Trump's negative sentiments:} Controversial scribe/remarks, undermining leadership image.
\\ \textbf{Harris' cites:} Focusing on justice, unity, and social growth.
\\ \textbf{Trump's cites:} Strong leadership, economic robustness.
\\ \textbf{Harris' main narratives:} Social justice, inclusivity, governance foresight.
\\ \textbf{Trump's main narratives:} Continuity of economic strength, sovereignty themes.
\\ \textbf{Favorite candidate summary:} Harris edges out due to strategic endorsements and key swing state focus.
\subsubsection{Dates: 2024-09-16 -- 2024-09-30}
 \textbf{Summary:} Harris and Trump maintain tight competition in the 2024 presidential race, engaging in strategic debates over issues such as immigration, economic policy, and foreign affairs. Harris sees growth in endorsements from party deferrers, while Trump's base remains resilient amidst controversies and legal pressures. Both candidates are evenly matched across crucial battleground states, turning debates and media encounters into imperative campaign tools.
\\ \textbf{Harris' probability score:} 0.5
\\ \textbf{Trump's probability score:} 0.5
\\ \textbf{Harris' positive sentiments:} Robust debate preparation and endorsements.
\\ \textbf{Trump's positive sentiments:} Economic security focus and strong partisan support.
\\ \textbf{Harris' negative sentiments:} Criticism about policy execution and border plans.
\\ \textbf{Trump's negative sentiments:} Rhetoric issues, questions surrounding fitness.
\\ \textbf{Harris' cites:} Highlighting democratic integrity.
\\ \textbf{Trump's cites:} Promising economic recovery.
\\ \textbf{Harris' main narratives:} Reforming economy, advancing social policies.
\\ \textbf{Trump's main narratives:} Highlighting economic achievements and strong foreign policy.
\\ \textbf{Favorite candidate summary:} Neither candidate emerges as a clear frontrunner given equal campaign leveraging on major elector matters.
\subsubsection{Dates: 2024-10-01 -- 2024-10-15}
 \textbf{Summary:} As election day approaches, the race between Harris and Trump intensifies with both engaging in strategic public appearances and emphasizing contrasting economic and social narratives. Harris showcases transparency and democratic values, leading in some polls, while Trump retains solid support through economic promises and a forceful personal campaign style. The debate over transparency, health, and policy details remain focal in shaping public opinion.
\\ \textbf{Harris' probability score:} 0.52
\\ \textbf{Trump's probability score:} 0.48
\\ \textbf{Harris' positive sentiments:} Transparency and bipartisan endorsements.
\\ \textbf{Trump's positive sentiments:} Vigor in campaigning and policy emphasizing.
\\ \textbf{Harris' negative sentiments:} Linked criticisms to ongoing policies.
\\ \textbf{Trump's negative sentiments:} Transparency concerns and controversial statements.
\\ \textbf{Harris' cites:} Protecting democracy and public welfare.
\\ \textbf{Trump's cites:} Addressing national economy and security.
\\ \textbf{Harris' main narratives:} Unity, social justice, and proactive governance.
\\ \textbf{Trump's main narratives:} Economic revival and direct leadership.
\\ \textbf{Favorite candidate summary:} Harris slightly favored due to poll momentum and broader appeal across key demographics.
\subsection{Results grouped by web resources}
\subsubsection{Web resource: Web sites}
 \textbf{Summary:} General resources depict the election battle between Harris and Trump as a competitive event with varying influences like media personalities and political endorsements. Issues ranging from economic policy to healthcare form core discussions around candidate capabilities. Harris receives affirmations from both expected and unexpected sectors, while Trump's steadfast support highlights his enduring appeal in specific demographics.
\\ \textbf{Harris' positive sentiments:} Dynamic leadership depiction with diverse endorsements.
\\ \textbf{Trump's positive sentiments:} Strong voter engagement and loyal demographic reach.
\\ \textbf{Harris' negative sentiments:} Concerns regarding issue clarity and youthful outreach.
\\ \textbf{Trump's negative sentiments:} Divisive remarks and portrayal mishandling.
\\ \textbf{Harris' cites:} Paramount in delivering justice and equity.
\\ \textbf{Trump's cites:} Economic achievements and self-presentation.
\\ \textbf{Harris' main narratives:} Adapting diversity, economic security, and social reform.
\\ \textbf{Trump's main narratives:} Resilience through economic frames and direct policy.
\subsubsection{Web resource: The New York Times}
 \textbf{Summary:} The New York Times discusses detailed insights involving political strategy, media effectiveness, and personalized voter outreach. Both candidates utilize diverse outreach platforms to strengthen campaign narratives, with public opinion swaying based on critical incident analysis (e.g., debate performance, endorsements). Harris focuses on recent political traction and narrative inclusivity, while Trump remains vital in reinforcing economic and governance issues crucial to his base.
\\ \textbf{Harris' positive sentiments:} Increased visibility and endorsement, wider electoral appeal.
\\ \textbf{Trump's positive sentiments:} Rich engagement in consistent themes of economic highlights.
\\ \textbf{Harris' negative sentiments:} Intersection between historical administration critiques and new mandates.
\\ \textbf{Trump's negative sentiments:} Challenges involving rhetoric and adherence to political narratives.
\\ \textbf{Harris' cites:} Appealing narrative inclusions fortifying democratic structures.
\\ \textbf{Trump's cites:} Strategizing around economic revival initiatives.
\\ \textbf{Harris' main narratives:} Engagement and reinforcing inclusivity.
\\ \textbf{Trump's main narratives:} Economic verification, school of resilience.
\subsubsection{Web resource: CNN}
 \textbf{Summary:} CNN analyses and reports showcase a politically charged environment where Harris garners support from statements and debates while Trump relies on established rhetoric around stability and economic prioritization. Both engage in media-fueled contests of leadership validation and public engagement, presenting near equal electoral prospects at this stage.
\\ \textbf{Harris' positive sentiments:} Democratic integrity advocacy, media appearance strength.
\\ \textbf{Trump's positive sentiments:} Economic and authoritative dominance emphasized.
\\ \textbf{Harris' negative sentiments:} Policy positioning evaluations and alternative handling.
\\ \textbf{Trump's negative sentiments:} Detractors identifying inconsistent statements and past legacy scrutiny.
\\ \textbf{Harris' cites:} Advocating societal reform and resolutions.
\\ \textbf{Trump's cites:} Economics-centered dialogues.
\\ \textbf{Harris' main narratives:} Pushing democratic restoration and societal reform.
\\ \textbf{Trump's main narratives:} Centre of stabilizing economic interests.
\subsubsection{Web resource: The Washington Post}
 \textbf{Summary:} The Washington Post reflects consistent campaign-centered dialogue with candidate differences ranging from economic management to border policy focus. Harris benefits from dual-party endorsements boosting profile themes, while Trump aggressively campaigned on economic security. Competitive terms revolve from strategic appeal to core contrasts in social values perception.
\\ \textbf{Harris' positive sentiments:} Enrichment from growing cross-supportive endorsements.
\\ \textbf{Trump's positive sentiments:} Strong baseline support and economic focal viewpoints.
\\ \textbf{Harris' negative sentiments:} Performance critiques highlighting administration engagement.
\\ \textbf{Trump's negative sentiments:} According perceptions of doubt primarily through rhetoric review.
\\ \textbf{Harris' cites:} Successfully engaging in political collaborations.
\\ \textbf{Trump's cites:} Projecting economic dichotomies or statements.
\\ \textbf{Harris' main narratives:} Strengthening bridges in diverse party coalition ethos.
\\ \textbf{Trump's main narratives:} Economic endurance and stability themes.
\subsubsection{Web resource: Fox News}
 \textbf{Summary:} Fox News navigates the political complexities with a focus on electoral system confidence and candidate speech engagement. Both Harris and Trump employ varied approaches across public appearances to address core policies and rally support. Harris drives forward on narratives depicting change while Trump sustains political traditionalism frameworks. Voter attention toward robust engagement and narrative reliability continues to dictate polling impacts.
\\ \textbf{Harris' positive sentiments:} Effective foundational change presentation.
\\ \textbf{Trump's positive sentiments:} Assurance in long-term economic stability traditions.
\\ \textbf{Harris' negative sentiments:} Skepticism faced regarding existing establishment and institutional relations.
\\ \textbf{Trump's negative sentiments:} Exaggerated position interpretations promoting dissent.
\\ \textbf{Harris' cites:} Narrative growth around the infusion of change policy initiatives.
\\ \textbf{Trump's cites:} Engagement with economic vitality longevity.
\\ \textbf{Harris' main narratives:} Essential to broadening change policy promotions.
\\ \textbf{Trump's main narratives:} Consistency themes across prior and ongoing economic implications.
\subsubsection{Web resource: Reuters}
 \textbf{Summary:} Reuters' insights highlight voter engagement meticulously carved through issue deep-dives including climate, economic revival with voter trust dependent on alignment shifts. Each candidate employs tailored rhetoric aimed at navigating pivotal swing outcomes. Public perception hinges on media impressions sustained by strategic and issue-focused campaign trail narratives.
\\ \textbf{Harris' positive sentiments:} Resilient social justice and reform commitment visibility.
\\ \textbf{Trump's positive sentiments:} Demographic connectivity in line with strong policy stances.
\\ \textbf{Harris' negative sentiments:} Exploration of intense policy outcome requirements.
\\ \textbf{Trump's negative sentiments:} Rhetoric stretched across controversial policy lines.
\\ \textbf{Harris' cites:} Policy-oriented informational sharing.
\\ \textbf{Trump's cites:} Slogans addressing core economic positions.
\\ \textbf{Harris' main narratives:} Continued drive for societal reform and economic innovation.
\\ \textbf{Trump's main narratives:} Thematic engagements emphasizing economic security bearings.
\subsubsection{Web resource: NBC News}
 \textbf{Summary:} NBC News evaluates the electoral contest with perspectives on integrity, transparency, and state engagement vital for verifying contested campaign positions. Candidates invoke strong public-facing narratives on a multitude always hitting core socio-political policies essential to voters. Trump echoes confidence and stability remarks while Harris advances contemporary remedies for challenges in the civic and social sphere.
\\ \textbf{Harris' positive sentiments:} Enhanced remedy and reform advocacy.
\\ \textbf{Trump's positive sentiments:} Core community confidence underpinning security measures.
\\ \textbf{Harris' negative sentiments:} Position relations linked to previous institutional leadership boundaries.
\\ \textbf{Trump's negative sentiments:} Adaptation pressures in processing flexibility regarding positional changes.
\\ \textbf{Harris' cites:} Societal development and problem solution initiatives.
\\ \textbf{Trump's cites:} Security interventions and core valuations.
\\ \textbf{Harris' main narratives:} Formative change advocacy for the social pillar.
\\ \textbf{Trump's main narratives:} Upholding stability and identifying supportive demographic concentrations.
\subsubsection{Web resource: Bloomberg}
 \textbf{Summary:} Bloomberg's in-depth campaign reporting focuses on strategic fundraising, key endorsements, and critical voter affinities in the Harris-Trump electoral face-off. Identifying newsmaker influence and economic priority in decisively impacting electoral bases is pivotal. Advanced socio-economic ties structured through clarification of strategic legacies define disrupt-or-support trajectories for both contenders.
\\ \textbf{Harris' positive sentiments:} Policy-driven fundraising surge and voter envelopment.
\\ \textbf{Trump's positive sentiments:} National economic assurance; fortified tenure affirmations.
\\ \textbf{Harris' negative sentiments:} Complexity interactions within new and recurring endorsement structures.
\\ \textbf{Trump's negative sentiments:} Long-standing alignment surfaces controversial resonances.
\\ \textbf{Harris' cites:} Community-focused fundraising advantage narratives.
\\ \textbf{Trump's cites:} Resilient economic outreach perspectives.
\\ \textbf{Harris' main narratives:} Infrastructure-centered equity and innovative civic influence.
\\ \textbf{Trump's main narratives:} Tenure continuity and traditional economic reform platforms.
\subsection{Trend summary} 
 \textbf{Summary:} Over time, the race between Kamala Harris and Donald Trump showcases fluctuating leads across demographic and battleground sectors, reflecting the substantial environmental shifts in campaign emphasis. Poll and voter sentiment analysis highlight uncertainties attributed to external influences and strategic decisions from each campaign.
\\ \textbf{Harris' trend summary:} Harris trends positively with diverse faction gains, leveraging strategic endorsements and voter engagement techniques to reinforce democratic stances while grappling with policy continuity divergences.
\\ \textbf{Trump's trend summary:} Trump slowly addressing economic strengths and stability concerns while confronting rhetoric critiques and incongruities, balancing core support with broader voter appeal.
\\ \textbf{Harris' main narratives:} Advocating foundational social developments together with broad policy connection frameworks.
\\ \textbf{Trump' main narratives:} Economic emphasis with prevailing regulations ensuring an enduring voter connection.
\\ \textbf{Favorite candidate's summary:} In summation, current trends favor Harris slightly based on strategic voter alignment, dynamic rallying figures, and committed ideological rapport widening her electoral reach.


\section{Quantitative Results}
\subsection{Distributions of quantitative scores}
\begin{figure}[H]
\center
 \includegraphics[width=1\linewidth]{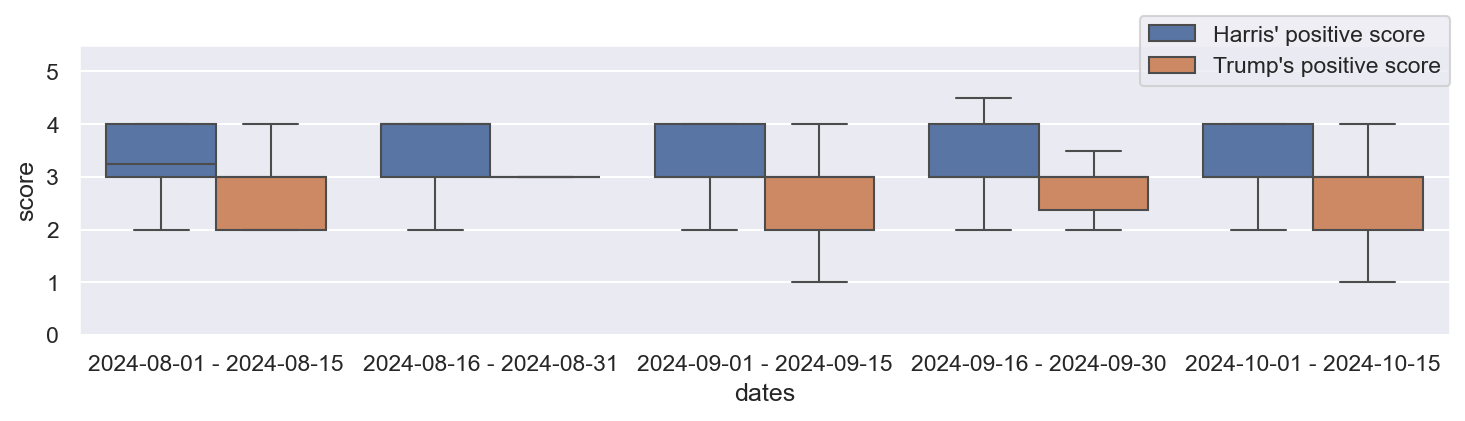}
 \caption{Candidates' positive scores for time periods}
 \label{positive_score}
 \end{figure}

\begin{figure}[H]
\center
 \includegraphics[width=1\linewidth]{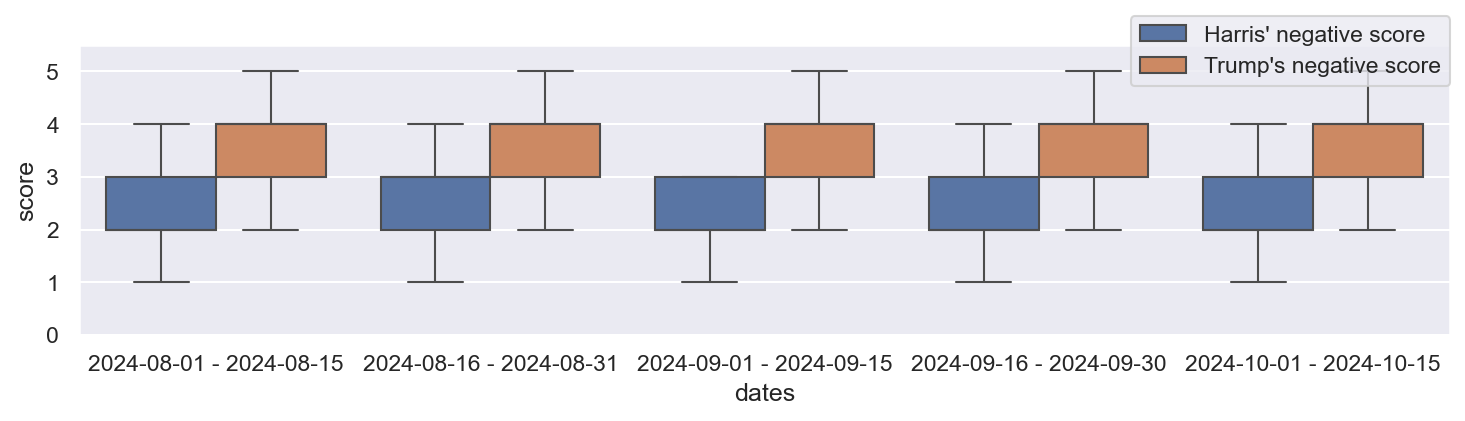}
 \caption{Candidates' negative scores for time periods}
 \label{negative_score}
 \end{figure}

\begin{figure}
\center
\includegraphics[width=1\linewidth]{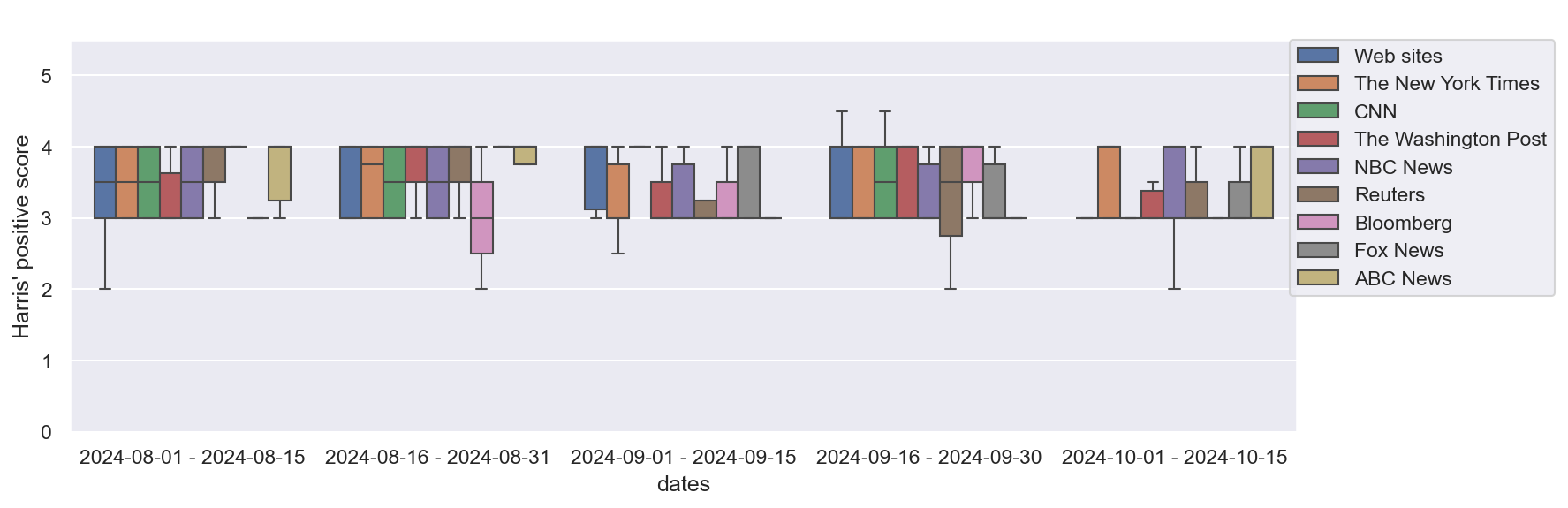}
\caption{Harris' positive sentiments  scores in web resources}
\label{webresource_harris_pos}
\end{figure}

\begin{figure}
\center
\includegraphics[width=1\linewidth]{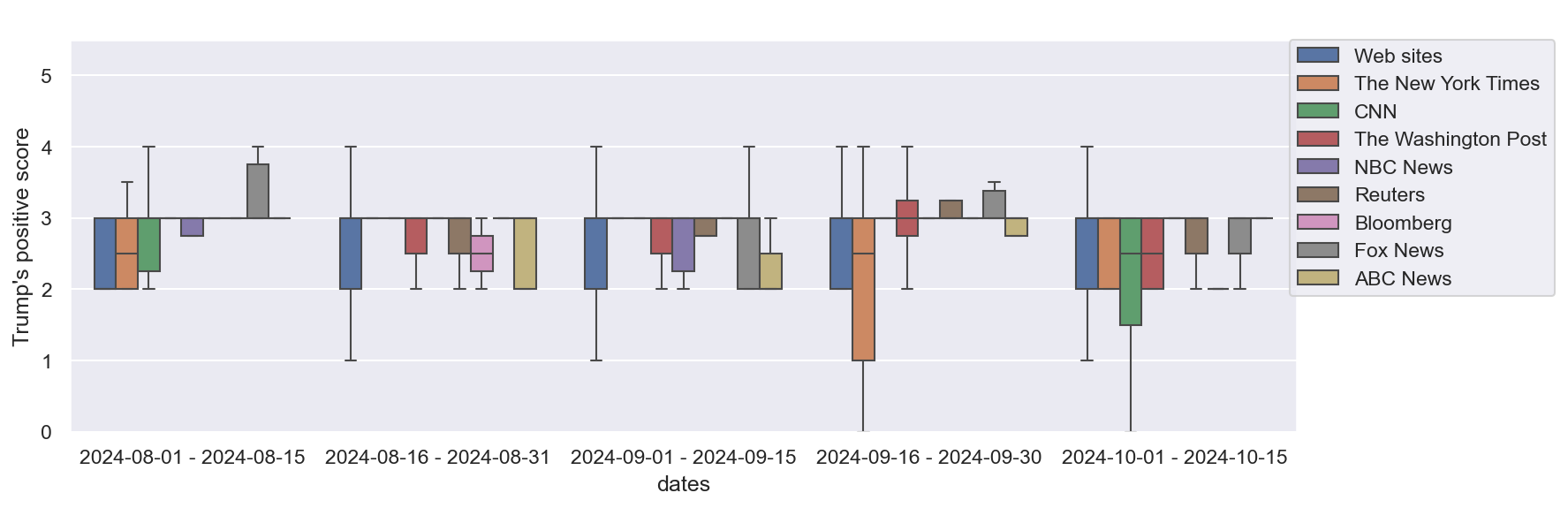}
\caption{Trump's positive sentiments  scores in web resources}
\label{webresource_trump_pos}
\end{figure}

\begin{figure}
\center
\includegraphics[width=1\linewidth]{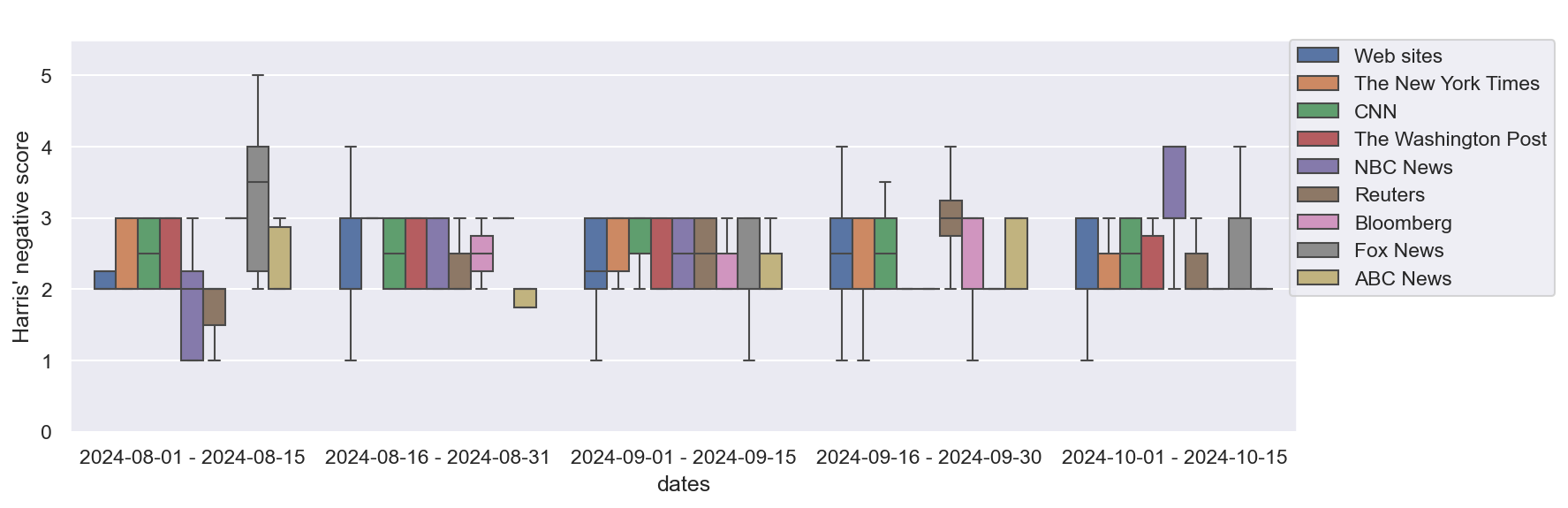}
\caption{Harris' negative sentiments  scores in web resources}
\label{webresource_harris_neg}
\end{figure}

\begin{figure}
\center
\includegraphics[width=1\linewidth]{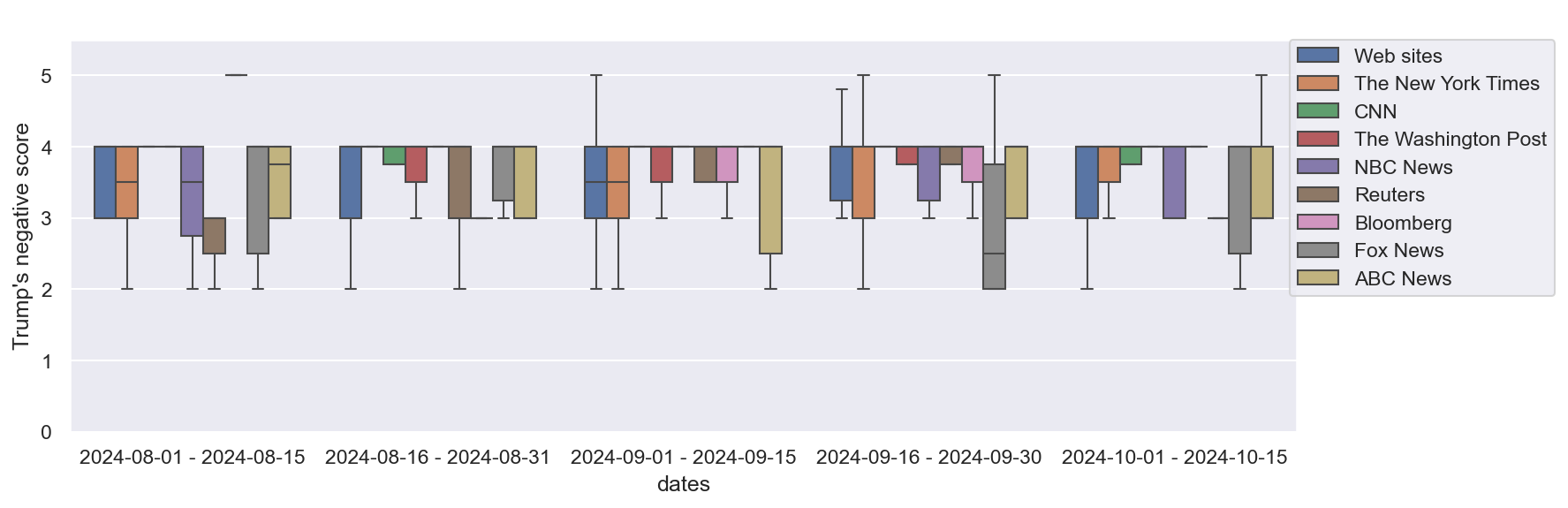}
\caption{Trump's negative sentiments  scores in web resources}
\label{webresource_trump_neg}
\end{figure}

\begin{figure}
\center
 \includegraphics[width=1\linewidth]{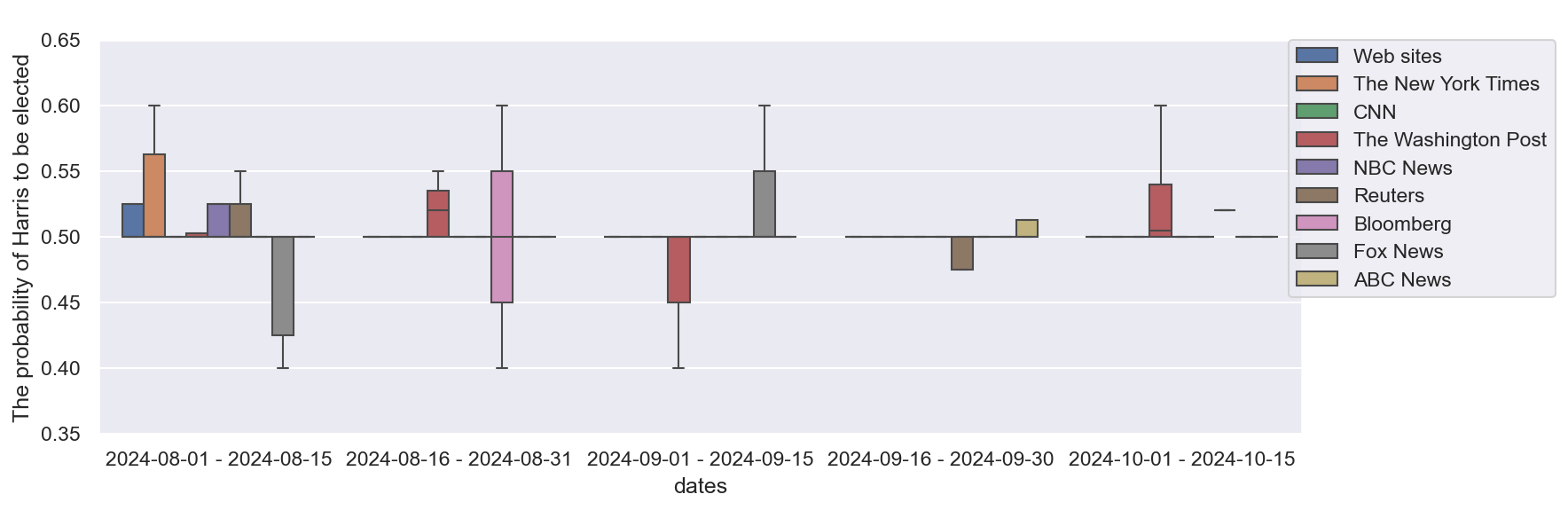}
 \caption{Probability for Harris to be elected in different web resources}
 \label{webresource_harris_prob_elect}
 \end{figure}

\begin{figure}
\center
 \includegraphics[width=1\linewidth]{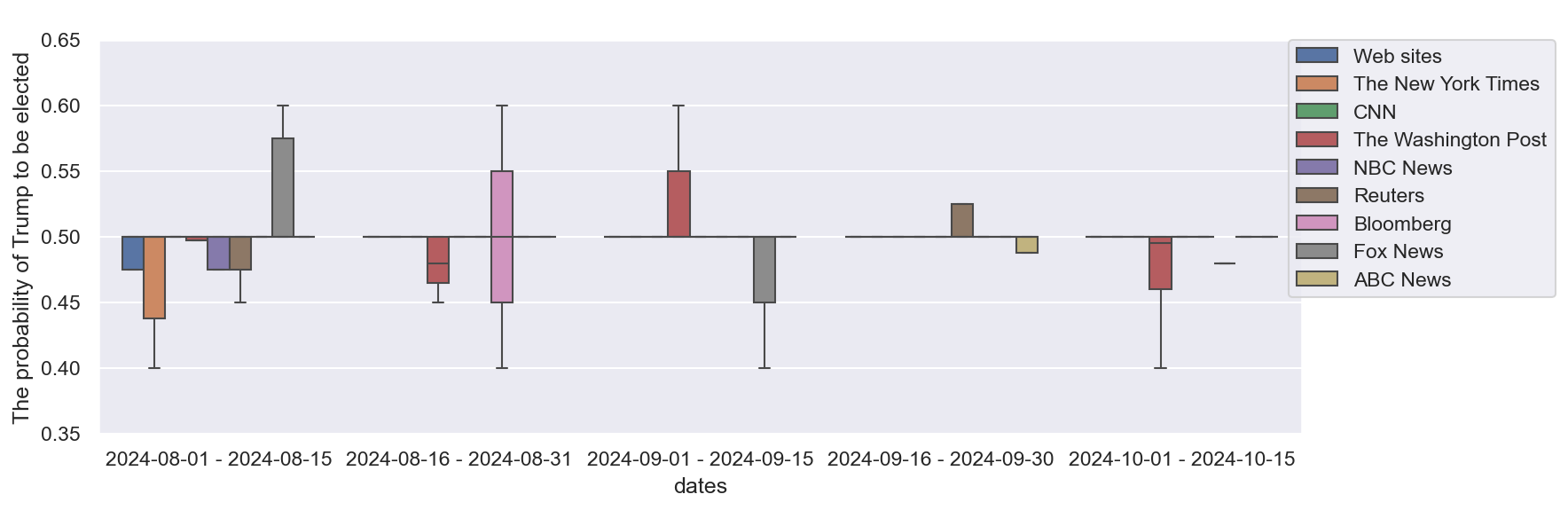}
 \caption{Probability for Trump to be elected in different web resources}
 \label{webresource_trump_prob_elect}
 \end{figure}

\begin{figure}
\center
 \includegraphics[width=0.85\linewidth]{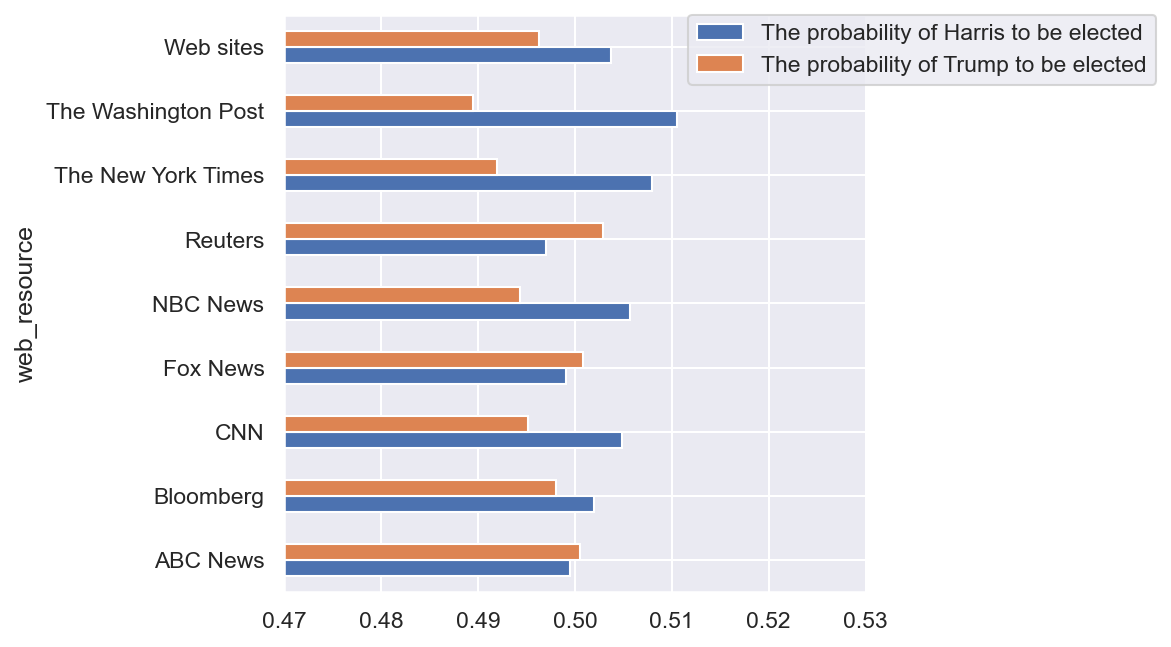}
 \caption{Mean values for the probability of candidates to be elected in web resources}
 \label{prob_Harris_Trump}
 \end{figure}

On the first level of RAG approach, GPT-4o model generated scores for  positive and negative sentiments  as well as probability to be elected for each candidates. The distribution of  sentiment quantitative scores in different time periods are presented by boxplots on Fig.~\ref{positive_score}--~\ref{negative_score}.
The similar distribution grouped by web resources are presented on Fig.~\ref{webresource_harris_pos}--~\ref{webresource_trump_neg}.  Probabilities of candidates to be elected are presented by boxplots on Fig.~\ref{webresource_harris_prob_elect}--\ref{webresource_trump_prob_elect}.
Fig.~\ref{prob_Harris_Trump} shows the mean values of probabilities of candidates to be elected for different web resources. As one can see these probability practically equal, at the same time sentiment scores have more high volatility and are different for both candidates.

\FloatBarrier
 
\FloatBarrier
 
\FloatBarrier

\subsection{Bayesian regression for score trends}
\begin{figure}
\center
\includegraphics[width=1\linewidth]{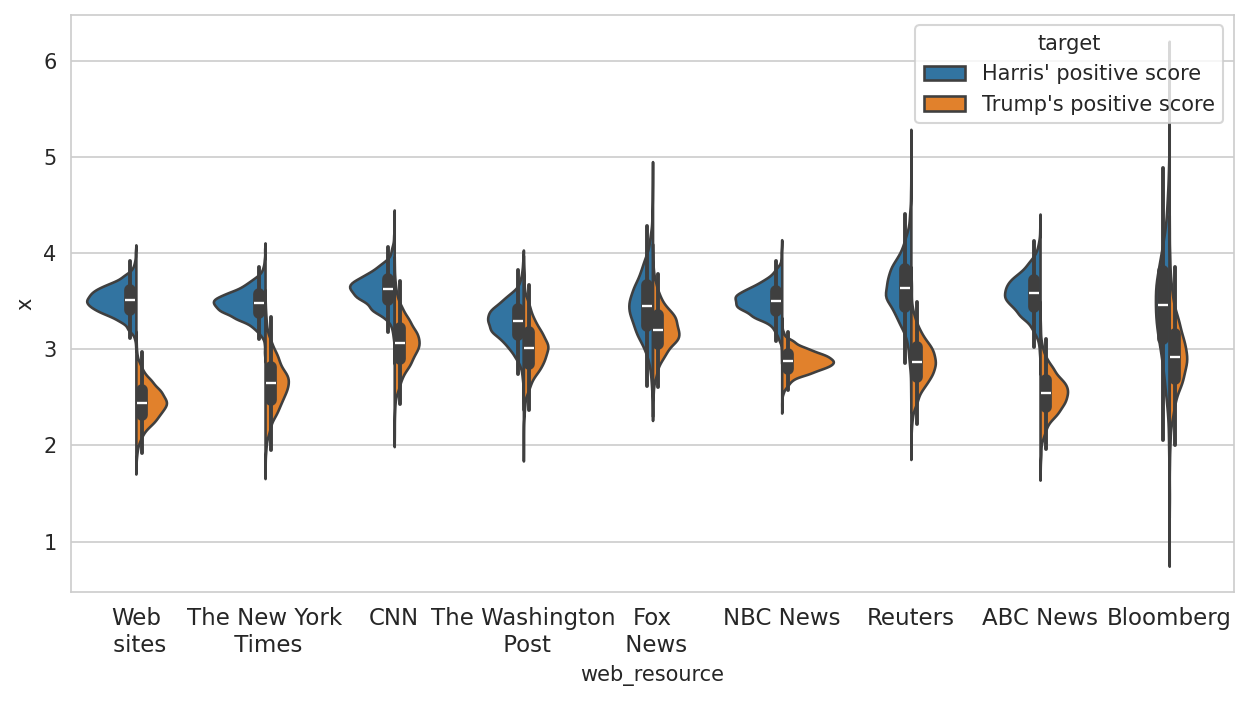}
\caption{Distribution for $\alpha$ parameter for candidates' positive sentiment score trend}
\label{positive_score_alpha}
\end{figure}

\begin{figure}
\center
\includegraphics[width=1\linewidth]{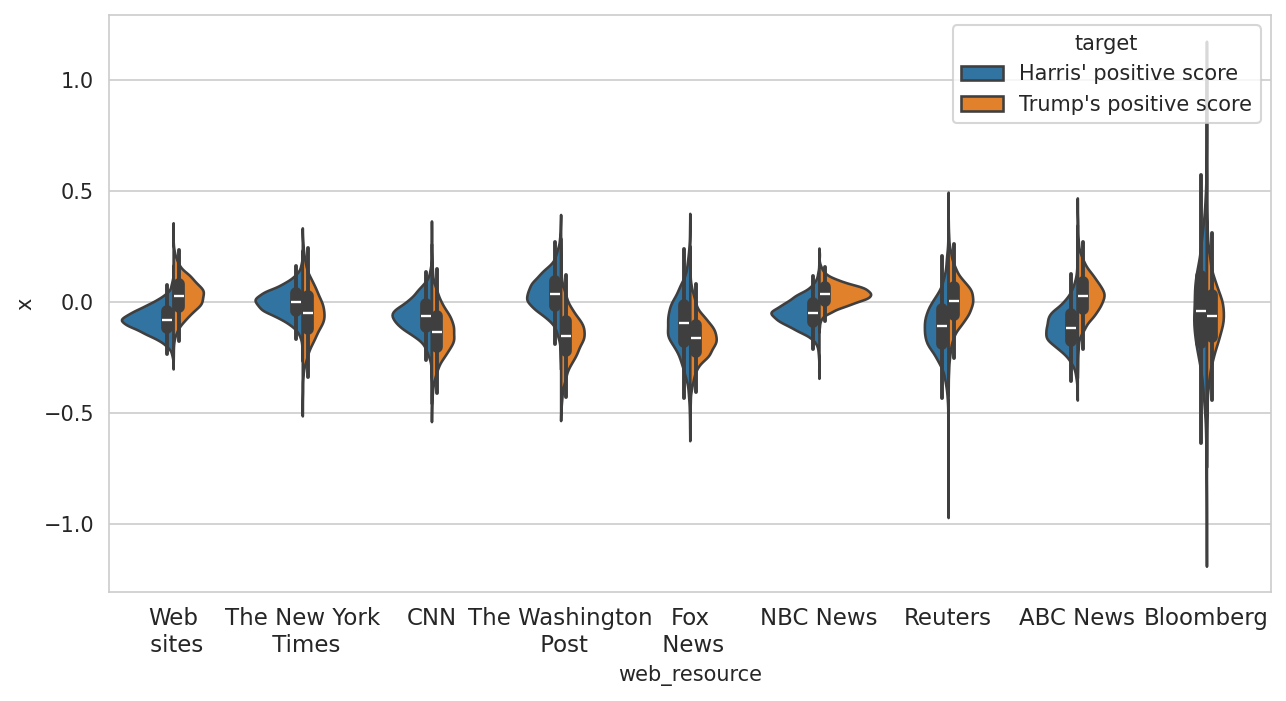}
\caption{Distribution for $\beta$ parameter for candidates' positive sentiment score trend}
\label{positive_score_beta}
\end{figure}

\begin{figure}
\center
\includegraphics[width=1\linewidth]{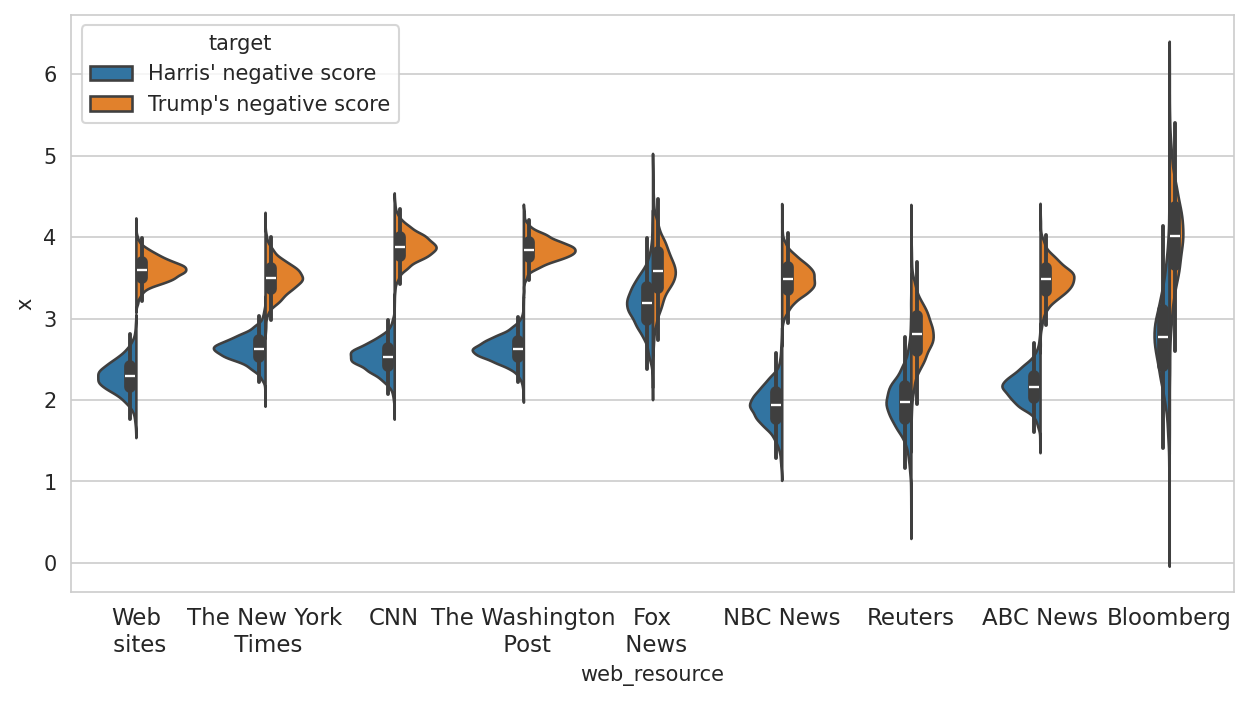}
\caption{Distribution for $\alpha$ parameter for candidates' negative sentiment score trend}
\label{negative_score_alpha}
\end{figure}

\begin{figure}
\center
\includegraphics[width=1\linewidth]{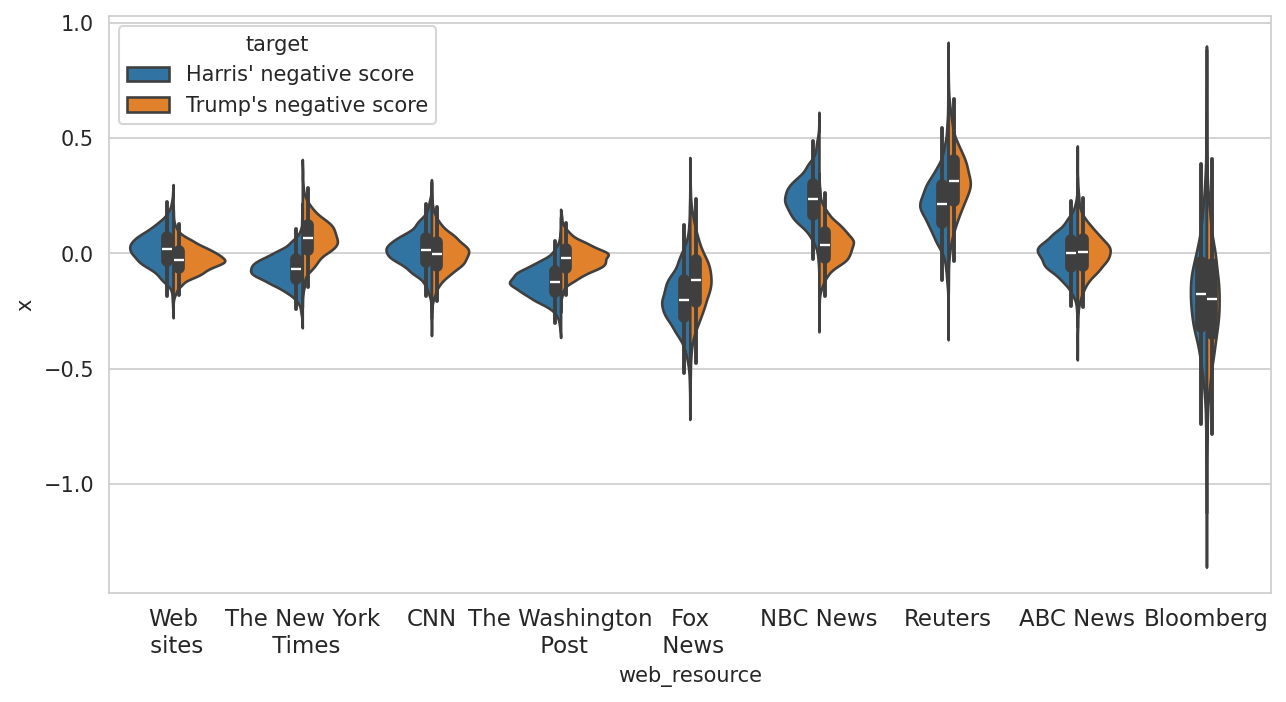}
\caption{Distribution for $\beta$ parameter for candidates' negative sentiment score trend}
\label{negative_score_beta}
\end{figure}

\begin{figure}
\center
\includegraphics[width=1\linewidth]{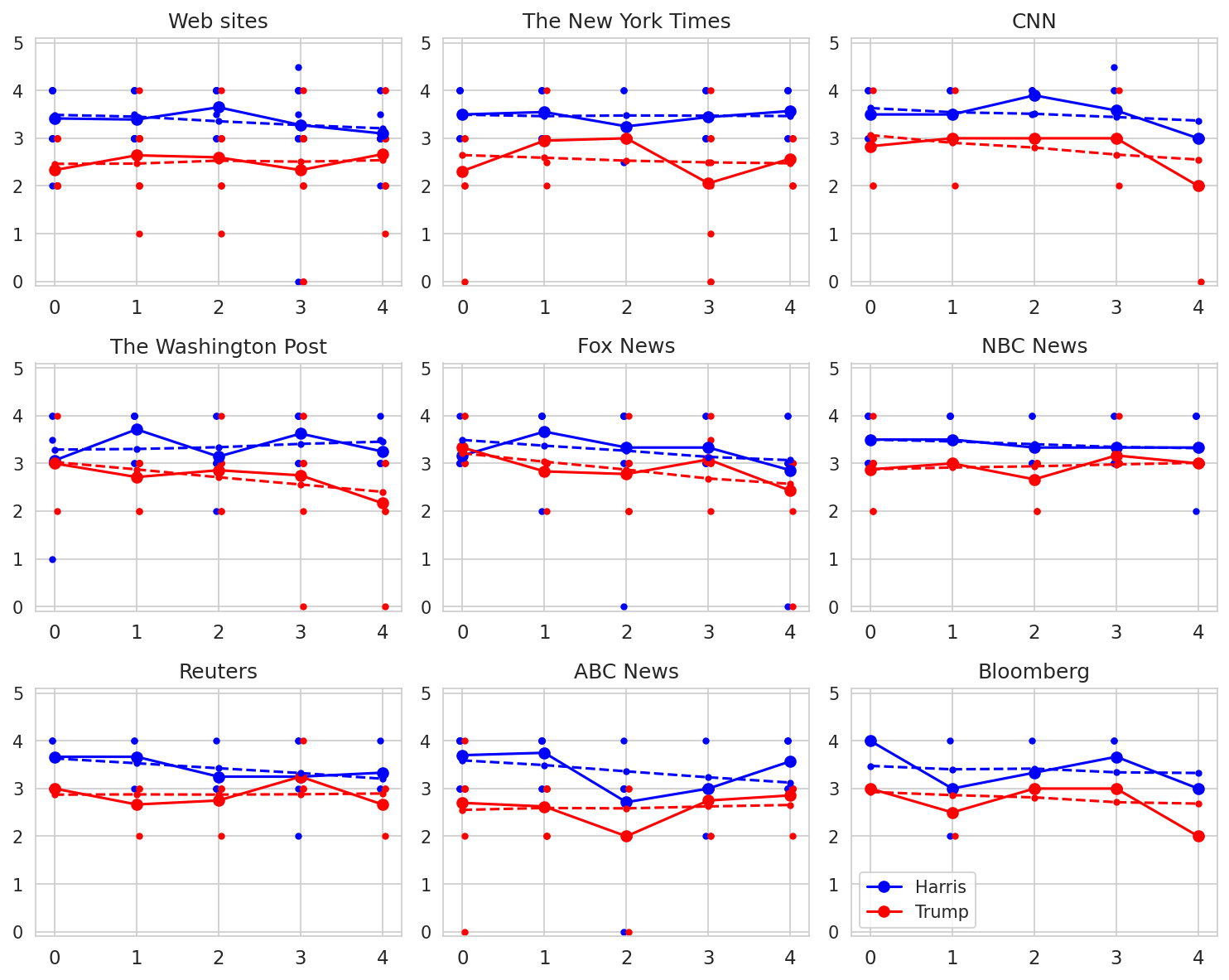}
\caption{Positive sentiment score trends for web resources}
\label{positive_score_trend}
\end{figure}

\begin{figure}
\center
\includegraphics[width=1\linewidth]{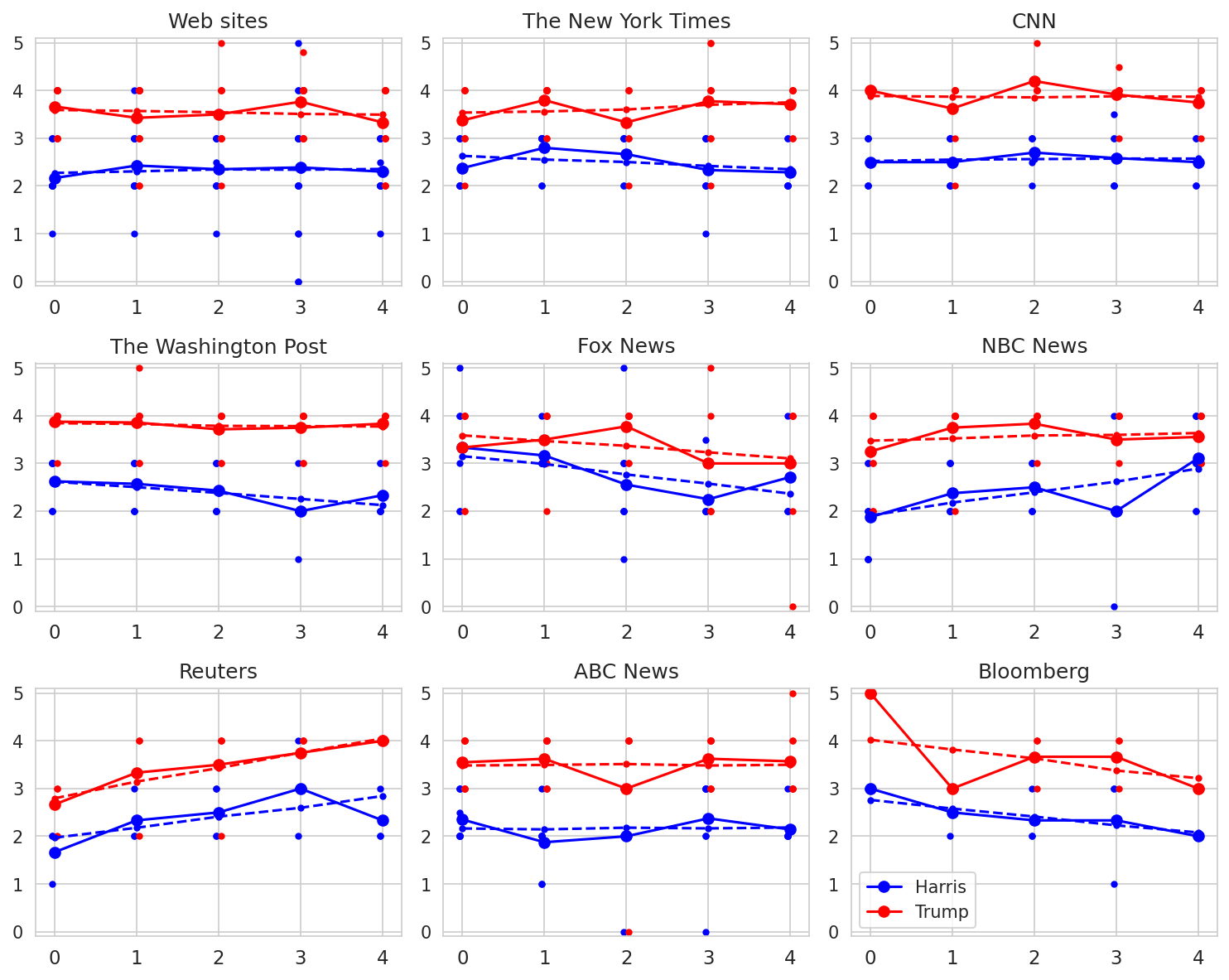}
\caption{Negative sentiment score trends for web resources}
\label{negative_score_trend}
\end{figure}

To analyze trends, we can use Bayesian regression~\cite{kruschke2014doing,gelman2013bayesian,carpenter2017stan}.
 This approach allows us to receive a posterior distribution of model parameters by using conditional likelihood and prior distribution~\cite{pavlyshenko2020using, pavlyshenko2020regression}. 
 The probabilistic approach makes it possible to receive the probability density function for the target variable. 
To analyze trends for sentiment scores, we used a Bayesian regression model as follows:
\begin{equation}
 \begin{split}
 &Score \sim \mathcal{N}(\mu, \sigma) \\
 &\mu=\alpha + \beta t 
 \end{split}
\end{equation}
where $\alpha$ parameter describes the bias (intercept) of the trend, which can be treated as the initial score when $t=0$ in the time period under consideration, $\beta$ parameter describes the slope for upward or downward score trends.
To solve Bayesian models, numerical Monte-Carlo methods are used. Gibbs and Hamiltonian sampling are popular methods for finding posterior distributions of the parameters in probabilistic models
~\cite{kruschke2014doing,gelman2013bayesian, carpenter2017stan}.
Bayesian inference makes it possible to obtain probability density functions for model parameters and estimate the uncertainty which is important in risk assessment analytics. For Bayesian inference calculations, we used \textit{PyStan} package for \textit{Stan} platform for statistical modeling \cite{carpenter2017stan}.
For time independent variables, we used a list of indexes for time periods [0,1,2,3,4] which correspond to the following time periods:'2024-08-01'--'2024-08-15', '2024-08-16'--'2024-08-31', '2024-09-01'--'2024-09-15', '2024-09-16'--'2024-09-30', '2024-10-01'--'2024-10-15'. 
 Boxplots for probability distributions of 
$\alpha$ and $\beta$ parameters for sentiment trends of both candidates are shown in Fig.~\ref{positive_score_alpha}--\ref{negative_score_beta}.
Fig.~\ref{positive_score_trend}--\ref{negative_score_trend} show the trends for different time periods and web resources. 
Points on the figures represent sentiment scores for specified time periods,  lines connect the mean values of the sentiment scores, and the dashed line shows the mean values of the Bayesian regression for the points of specified time periods.

\FloatBarrier

\section{Conclusion}
In this study, we consider the approach of using Google Search API and GPT-4o model for qualitative and quantitative analysis of news through retrieval-augmented generation (RAG). This approach was applied to analyse  news about the 2024 U.S. presidential election process. Different news sources for different time periods were analyzed. Quantitative sentiment scores generated by the GPT model were analyzed using Bayesian regression to get trend lines. The distributions found for the regression parameters allow for the analysis an uncertainty in the election process. 
The approach does not aim to predict the election outcome, as it does not take into account  the specific features of the U.S. election system, nor does it analyze the geospatial and state structure of quantitative scores. However, it can be used as part of a complex analytical approach. 
The results show that probabilities to be elected for both candidates are very similar, despite differences in their sentiment scores.
One of the main goals of this study is to provide qualitative and quantitative analytical information to political  news experts for further analysis. 
The obtained results demonstrate that using a GPT models for news analysis  can yield informative qualitative and quantitative analytics, providing important insights which can be used in the next stages of presidential election process analytics. 

\section{Disclaimer} 
The approach, ideas, and results shared in this study are for academic purposes only and are not intended to inform real-world conclusions or recommendations.

\bibliographystyle{unsrt}
\bibliography{article.bib}

\begin{thebibliography}{10}

\bibitem{achiam2023gpt}
Josh Achiam, Steven Adler, Sandhini Agarwal, Lama Ahmad, Ilge Akkaya,
  Florencia~Leoni Aleman, Diogo Almeida, Janko Altenschmidt, Sam Altman,
  Shyamal Anadkat, et~al.
\newblock {GPT-4 Technical Report}.
\newblock {\em arXiv preprint arXiv:2303.08774}, 2023.

\bibitem{pavlyshenko2023financial}
Bohdan~M Pavlyshenko.
\newblock {Financial News Analytics Using Fine-Tuned Llama 2 GPT Model}.
\newblock {\em arXiv preprint arXiv:2308.13032}, 2023.

\bibitem{touvron2023llama}
Hugo Touvron, Louis Martin, Kevin Stone, Peter Albert, Amjad Almahairi, Yasmine
  Babaei, Nikolay Bashlykov, Soumya Batra, Prajjwal Bhargava, Shruti Bhosale,
  et~al.
\newblock {Llama 2: Open foundation and fine-tuned chat models}.
\newblock {\em arXiv preprint arXiv:2307.09288}, 2023.

\bibitem{pavlyshenko2023analysis}
Bohdan~M Pavlyshenko.
\newblock {Analysis of Disinformation and Fake News Detection Using Fine-Tuned
  Large Language Model}.
\newblock {\em arXiv preprint arXiv:2309.04704}, 2023.

\bibitem{pavlyshenko2022methods}
Bohdan~M. Pavlyshenko.
\newblock {Methods of Informational Trends Analytics and Fake News Detection on
  Twitter}.
\newblock {\em arXiv preprint arXiv:2204.04891, Download PDF:
  https://arxiv.org/pdf/2204.04891.pdf}, 2022.

\bibitem{kawintiranon2022polibertweet}
Kornraphop Kawintiranon and Lisa Singh.
\newblock {PoliBERTweet: a pre-trained language model for analyzing political
  content on Twitter}.
\newblock In {\em Proceedings of the Thirteenth Language Resources and
  Evaluation Conference}, pages 7360--7367, 2022.

\bibitem{argyle2023out}
Lisa~P Argyle, Ethan~C Busby, Nancy Fulda, Joshua~R Gubler, Christopher
  Rytting, and David Wingate.
\newblock {Out of one, many: Using language models to simulate human samples}.
\newblock {\em Political Analysis}, 31(3):337--351, 2023.

\bibitem{goyal2022news}
Tanya Goyal, Junyi~Jessy Li, and Greg Durrett.
\newblock {News summarization and evaluation in the era of GPT-3}.
\newblock {\em arXiv preprint arXiv:2209.12356}, 2022.

\bibitem{rodman2024political}
Emma Rodman.
\newblock {On political theory and large language models}.
\newblock {\em Political Theory}, 52(4):548--580, 2024.

\bibitem{luitse2021great}
Dieuwertje Luitse and Wiebke Denkena.
\newblock {The great transformer: Examining the role of large language models
  in the political economy of AI}.
\newblock {\em Big Data \& Society}, 8(2):20539517211047734, 2021.

\bibitem{kheiri2023sentimentgpt}
Kiana Kheiri and Hamid Karimi.
\newblock {Sentimentgpt: Exploiting gpt for advanced sentiment analysis and its
  departure from current machine learning}.
\newblock {\em arXiv preprint arXiv:2307.10234}, 2023.

\bibitem{jiang2024disinformation}
Bohan Jiang, Zhen Tan, Ayushi Nirmal, and Huan Liu.
\newblock {Disinformation detection: An evolving challenge in the age of llms}.
\newblock In {\em Proceedings of the 2024 SIAM International Conference on Data
  Mining (SDM)}, pages 427--435. SIAM, 2024.

\bibitem{rozado2023political}
David Rozado.
\newblock {The political biases of chatgpt}.
\newblock {\em Social Sciences}, 12(3):148, 2023.

\bibitem{rathje2024gpt}
Steve Rathje, Dan-Mircea Mirea, Ilia Sucholutsky, Raja Marjieh, Claire~E
  Robertson, and Jay~J Van~Bavel.
\newblock {GPT is an effective tool for multilingual psychological text
  analysis}.
\newblock {\em Proceedings of the National Academy of Sciences},
  121(34):e2308950121, 2024.

\bibitem{spitale2023ai}
Giovanni Spitale, Nikola Biller-Andorno, and Federico Germani.
\newblock {AI model GPT-3 (dis) informs us better than humans}.
\newblock {\em Science Advances}, 9(26):eadh1850, 2023.

\bibitem{kruschke2014doing}
John Kruschke.
\newblock {\em {Doing Bayesian data analysis: A tutorial with R, JAGS, and
  Stan}}.
\newblock Academic Press, 2014.

\bibitem{gelman2013bayesian}
Andrew Gelman, John~B Carlin, Hal~S Stern, David~B Dunson, Aki Vehtari, and
  Donald~B Rubin.
\newblock {\em {Bayesian data analysis}}.
\newblock Chapman and Hall/CRC, 2013.

\bibitem{carpenter2017stan}
Bob Carpenter, Andrew Gelman, Matthew~D Hoffman, Daniel Lee, Ben Goodrich,
  Michael Betancourt, Marcus Brubaker, Jiqiang Guo, Peter Li, and Allen
  Riddell.
\newblock {Stan: A probabilistic programming language}.
\newblock {\em Journal of statistical software}, 76(1), 2017.

\bibitem{pavlyshenko2020using}
Bohdan~M. Pavlyshenko.
\newblock {Using Bayesian regression for stacking time series predictive
  models}.
\newblock In {\em 2020 IEEE Third International Conference on Data Stream
  Mining \& Processing (DSMP). Download PDF: https://arxiv.org/pdf/2201.02034},
  pages 305--309. IEEE, 2020.

\bibitem{pavlyshenko2020regression}
Bohdan~M. Pavlyshenko.
\newblock {Regression approach for modeling COVID-19 spread and its impact on
  stock market}.
\newblock {\em arXiv preprint arXiv:2004.01489. Download PDF:
  https://arxiv.org/pdf/2004.01489}, 2020.

\end{thebibliography}
\FloatBarrier
\end{document}